\documentclass[letterpaper]{article} 
\usepackage{aaai24}  
\usepackage{times}  
\usepackage{helvet}  
\usepackage{courier}  
\usepackage[hyphens]{url}  
\usepackage{graphicx} 
\usepackage{natbib}  
\usepackage{caption} 
\usepackage{algorithm}
\usepackage{algorithmic}

\usepackage{amssymb}
\usepackage{tabularx}
\usepackage{booktabs}
\usepackage{subfigure}
\usepackage{makecell}

\usepackage{newfloat}
\usepackage{listings}
\DeclareCaptionStyle{ruled}{labelfont=normalfont,labelsep=colon,strut=off} 
\floatstyle{ruled}
\newfloat{listing}{tb}{lst}{}
\floatname{listing}{Listing}
\title{Aligned with LLM: a new multi-modal training paradigm for encoding fMRI activity in visual cortex}
\author{
    Shuxiao Ma, Linyuan Wang, Senbao Hou, Bin Yan\textsuperscript{\rm }\thanks{Bin Yan is the corresponding author of this paper.}
}
\affiliations{
    \textsuperscript{\rm }Henan Key Laboratory of Imaging and Intelligent Processing\\

	PLA Strategic Support Force Information Engineering University\\
	Zhengzhou 450001, China\\
    \{mashuxiao23, wanglinyuanwly, hsb378093739\}@163.com, ybspace@hotmail.com
}

\usepackage{bibentry}

\begin{document}

\maketitle

\begin{abstract}
	
Recently, there has been a surge in the popularity of pre-trained large language models (LLMs) (such as GPT-4), sweeping across the entire Natural Language Processing (NLP) and Computer Vision (CV) communities. These LLMs have demonstrated advanced multi-modal understanding capabilities and showcased strong performance across various benchmarks. The LLM has started to embody traits of artificial general intelligence, which holds vital guidance for enhancing brain-like characteristics within visual encoding models. Hence, This paper proposes a new multi-modal training paradigm, aligning with LLM, for encoding fMRI activity in visual cortex. Based on this paradigm, we trained a encoding model in fMRI data named \textbf{LLM-}\textbf{V}isual \textbf{E}ncoding \textbf{M}odel (LLM-VEM). Specifically, we utilize LLM (miniGPT-4) to generate descriptive text for all stimulus images, forming a high-quality textual description set. Moreover, we use the pre-trained text encoder (CLIP) to process these detailed descriptions,obtaining the text embedding features. Next, we use the contrast loss function to minimize the distance between the image embedding features and the text embedding features to complete the alignment operation of the stimulus image and text information. With the assistance of the pre-trained LLM, this alignment process facilitates better learning of the visual encoding model, resulting in higher precision. The final experimental results indicate that our training paradigm has significantly aided in enhancing the performance of the visual encoding model.

\end{abstract}

\section{Introduction}

The way things happen or are experienced is termed as modality, which can be different sensory experiences or types of information, such as audio, video, images, and text \cite{MultimodalMachineLearning}. In recent years, the exponentially growing multi-modal data describing similar or related objects has made multi-modality a primary form of information resources \cite{du2007shape}. Our world is fundamentally composed of multiple coexisting modalities, and humans perceive the world through different senses, obtaining various modalities of information. The relationships between these different modalities are complementary and cooperative. Multi-modal machine learning methods resemble more closely the way humans perceive their surrounding environment, requiring an overarching architecture to understand, control, and coordinate modal information from different senses \cite{han2023survey}.

When humans acquire external information, the majority of it is obtained through vision. In other words, visual information becomes the primary source through which our brain receives external stimuli. Therefore, studying how the brain processes visual information is crucial for exploring and understanding the mechanisms of brain information processing. By constructing visual information encoding models in functional magnetic resonance imaging (fMRI) activity, researchers can investigate and predict the brain's responses to different visual stimuli. This modeling approach allows us to better comprehend how the brain encodes and processes visual information from perception to cognition. Visual information encoding models seek to simulate the information processing in the visual regions of the human brain. Their purpose is to predict the changes in responses across different voxels when exposed to various external stimulus images. \cite{1,2}.

\begin{figure*}[t]
	\centering
	\includegraphics[width=1.0\textwidth]{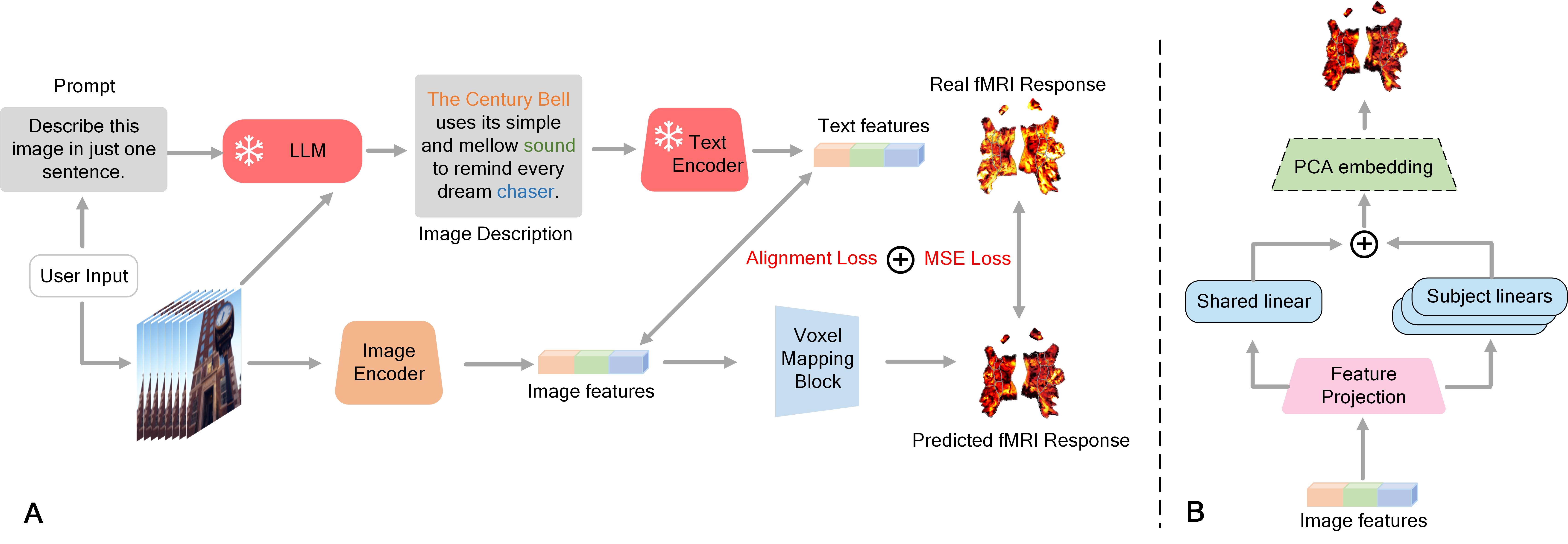} 
	\caption{The overall architecture of the proposed multi-modal training paradigm, aligning with LLM, for encoding fMRI activity in visual cortex. \textbf{A} represents the process of the training paradigm we designed; \textbf{B} illustrates the specific details of the Voxel Mapping Block in A.}
	\label{fig1}
\end{figure*}

In addition to containing a wealth of image feature information, visual information also comprises several other modalities of feature information. Ivanova \cite{ivanova2022role} discovered that during non-linguistic semantic tasks, such as image visual tasks, the language regions in the brain get activated. This indicates the involvement of language-related textual information in processing visual information. Previous studies have indicated that individuals with normal neural development may exhibit a phenomenon where the brain re-encodes the stimulus components of images into linguistic forms. This re-encoding process serves as an additional source of task-related information \cite{22, 23, 24}. These pieces of information collectively indicate a fact: when the brain processes visual information, apart from requiring a substantial amount of stimulus image features, appropriately integrating textual feature information is advantageous for visual information processing. Our external information exists in a multi-modal form, and the human brain's visual area also requires textual information as a secondary modality to assist in processing visual information. Building upon this theory, Liu et al. proposed the BrainCLIP model \cite{liu2023brainclip} that a key aspect of their work involves decoding visual information by combining visual and text supervision, embedding fMRI into the CLIP space. The success of their research confirms the significance of multi-modal information in the visual information processing flow, demonstrating its indispensability in the mechanism of visual information processing. Therefore, how to use textual feature information in visual encoding models to construct a multi-modal processing framework is a pressing issue awaiting resolution.

With the launch of ChatGPT in November 2022, representing GPT-4 \cite{38}, the artificial intelligence-generated content (AIGC) model sparked a significant wave within the community, showcasing extraordinary multimodal capabilities. It not only handles diverse text inputs but also comprehends the content of input images, generating detailed descriptions about them. ChatGPT has significantly propelled the development of artificial intelligence, altering the way humans interact with computers. Among these advancements, the rise of large language models (LLMs) like Flan-T5 \cite{chung2022scaling}, Vicuna \cite{vicuna2023}, LLaMA \cite{touvron2023llama}, and Alpaca \cite{taori2023stanford} has been particularly noteworthy. LLMs demonstrate robust abilities in context learning, reasoning, and comprehension, laying the groundwork for constructing the skyscraper of general artificial intelligence. However, the extraordinary capabilities of GPT-4 stem from its training on tens of billions of parameters across thousands of GPUs using millions of datasets. Faced with such high computational costs, it's nearly impossible for smaller communities to directly train their own GPT-4 level models. Moreover, the generation capabilities of chatbot models may not suffice for most practical production scenarios. For instance, traditional visual models like CNNs and ViTs remain the optimal choice for visual perception tasks such as image classification. Therefore, exploring how to leverage pre-trained LLMs to enhance the performance of existing task-specific models is more practical and cost-effective \cite{ding2023can}.

Trained deep neural networks (DNNs) aimed at solving computer vision perception tasks extract feature representations that accurately predict human responses to complex real-world visual stimuli, such as photos. Achieving this success required considerable effort from researchers. One hypothesis suggests that the predictive accuracy of DNNs depends on the hierarchical structure they employ for visual processing. This hypothesis is reasonable, as hierarchical processing is considered an important organizational principle in the visual systems of primates \cite{hubel1962receptive, riesenhuber2000computational, felleman1991distributed, richards2019deep, himberger2018principles}. In fact, some evolved DNN structures have been inspired by evidence of hierarchical processing in primate visual systems \cite{lecun1989backpropagation, krizhevsky2012imagenet, lecun2015deep}, and this hypothesis has received some experimental support. However, in the training paradigm of visual encoding models based on computer vision tasks, models only learn visual representations from stimulus images, disregarding information from other modalities. This single-modal training paradigm creates a significant gap between the processing flow of encoding models and the human brain's visual information processing flow.

To bridge this gap, this paper proposes a new multi-modal training paradigm, aligning with LLM, for encoding fMRI activity in visual cortex. This training paradigm leverages the knowledge of pre-trained LLMs to assist encoding models in learning enhanced image features from multi-modal sources, achieving higher performance. Specifically, we first employ the multi-modal understanding capability of pre-trained LLM to generate detailed high-quality descriptions for all stimulus images in the dataset. Subsequently, we utilize a pre-trained text encoder to extract corresponding text embeddings for all descriptions in the dataset. Through this process, the rich semantic information contained in the descriptive text generated by LLM is compressed into high-dimensional vectors, easily digestible by regular neural networks. Finally, during the training process, in addition to receiving supervised signals from fMRI, the model learns stimulus image features by minimizing a distance loss function aligned with text embeddings. By aligning these embeddings, the knowledge of image understanding obtained within LLM can be transferred to traditional visual encoding models across the image and text modalities.

Based on this training paradigm, we propose a model named the \textbf{LLM-}\textbf{V}isual \textbf{E}ncoding \textbf{M}odel (LLM-VEM). In LLM-VEM, the handling of stimulus image features involves a two-stage training process: Stage 1: In this phase, we utilize a frozen image feature extractor (EVA \cite{fang2023eva}) to perform feature extraction. Subsequently, we reduce the dimensionality of the image features through feature projection. Typically, the visual encoding model encounters over-fitting issues due to mismatches between dataset size and model parameters. To address this problem, we partially replace a portion of the voxel mapping network in the encoding model with a Principal Component Analysis (PCA) module. This step reduces the parameter count of the encoding model, alleviating over-fitting concerns. Stage 2: Using the best checkpoint saved from the first stage, we set several blocks within EVA to an unfrozen state while freezing other blocks of LLM-VEM. In this stage, we add the loss function aligned with LLM to align the stimulus image and text, enhancing the performance of our model. The overall design framework of the LLM-VEM model is illustrated in Fig. 1.

The main contributions of our work are as follows:

\begin{itemize}
	\item We have introduced, for the \textbf{first} time, a multi-modal training paradigm, aligning with LLM, for encoding fMRI activity in visual cortex.
	\item Based on this training paradigm, we propose a visual encoding model that aligns the embedded features of stimulus images with the text embeddings generated by LLM. Our model employs a contrastive learning loss function to minimize the distance between the stimulus image features and text features, aiming to achieve alignment between the image and text.
	\item We employed LLM to produce a high-quality textual description dataset for the NSD dataset, supplying a source of multimodal information for the multimodal representation of stimulus image features.
	
\end{itemize}

\section{Methods}

\subsection{Overall Architecture}

In this section, we initially present the overall architecture of the training paradigm. Subsequently, we discuss the process of generating image descriptions using multi-modal LLM. Finally, we explain how these descriptions serve as supervised signals during training, aiding the visual encoding model to acquire better feature representations, thereby achieving improved performance in subsequent visual encoding tasks.

The training paradigm aligned with LLM primarily consists of two stages: Stage 1 involves the initialization training of a conventional visual encoding model to obtain the best model checkpoint. It is noteworthy that in this phase, the model's feature extractor remains frozen, and the training is solely focused on the voxel mapping block that generates fMRI signals from the features. Stage 2 commences from the optimal checkpoint obtained in stage one, where most modules in the LLM-VEM model have their parameters frozen. Only a few blocks within the image feature extractor are unfrozen, contributing to the fine-tuning process of the model. During this fine-tuning of the stimulus image feature extractor, we incorporate the alignment of stimulus image features with LLM. Fig. 1 illustrates the overall architecture of the proposed multi-modal training paradigm, aligning with LLM, for encoding fMRI activity in visual cortex.

Subplot A in Fig. 1 encompasses the entire process of both stage 1 and stage 2 in our training paradigm. From a different perspective, after the model receives the stimulus image and the inquiry phrase, the LLM generates textual descriptive information about the stimulus image based on the input. This textual information then undergoes embedding via the text encoder, resulting in embedded textual feature information. Simultaneously, the stimulus image undergoes encoding through the image encoder, producing image feature information. The image features are then predicted through the voxel mapping component to estimate the corresponding fMRI signal. This predicted value is compared to the fMRI signal in our dataset using a minimal distance loss function to obtain the optimal predicted fMRI value. Meanwhile, through minimizing an alignment loss function between the textual and image features, the stimulus image features attempt to align with the LLM, achieving the multi-modal representation of the stimulus image. Subplot B in Figure 1 provides a detailed description of the components within the voxel mapping block. For further elaboration, please refer to the section titled "Voxel Mapping Block" in the following text.

\subsection{Image Extractor Block}

To enhance the encoding performance, in our study, we utilized the stimulus image extraction model of the pre-trained large-scale model EVA \cite{fang2023eva} in stage 1. EVA is a visual base model with billions of parameters. During training, EVA masks parts of its output to reconstruct the corresponding output of the CLIP model, thus enabling EVA to possess both strong semantic learning capabilities like CLIP and robust geometric structure learning abilities akin to MIM, in a simple and efficient manner.

\subsection{Voxel Mapping Block}

In the realm of human visual encoding and decoding, datasets commonly used often contain a limited number of pairs of stimulating images and fMRI samples. When dealing with DNNs that usually have a large number of parameters, visual encoding models are prone to over-fitting issues. To mitigate over-fitting problems, we primarily employed the following two measures: 1.Unified encoding strategy: We combined stimulating images and fMRI from multiple subjects in the dataset, forming a larger dataset. This approach expands the dataset size from a training perspective. 2. In the final stage of the voxel mapping module, instead of using DNNs, we utilized PCA (Principal Component Analysis) technique to accomplish the voxel mapping function. Specifically, we performed PCA on the datasets of all subjects, setting the weights and biases as the center and bias of the PCA, respectively.

\subsection{Generating high-quality textual descriptions}

In this section, we discussed how to use pre-trained generative language models (such as BLIP-2 \cite{li2023blip} and MiniGPT-4 \cite{zhu2023minigpt}) to generate textual descriptions of stimulus images. These models take stimulus images and user prompts as input and provide corresponding textual descriptions.

Initially, by using prompts like "Describe this image in one sentence," we can obtain a sentence-based description of the stimulus image. Although such descriptions may not cover all the details within the stimulus image, they capture the most salient information present in the image.

\subsection{Alignment process of stimulus images and textual information}

In the previous section, we generated textual descriptions of stimulating images using LLM. In this section, we will illustrate how to utilize these textual descriptions within the training process of a visual encoding model to enhance the features of stimulating images. This enhancement aims to improve the performance of the encoding model.

Generally, a visual encoding model comprises a stimulating image feature extraction model and a voxel mapping model. Using $F$ be a stimulating image feature extracting model which uses an image $im{{g}_{i}}$ as input, and the image feature is ${{f}_{i}}=F(im{{g}_{i}})$. Similarly, using $T$ be a voxel mapping model which uses an image feature ${{f}_{i}}$ as input, and the predicted voxel is ${{v}_{i}}=T({{f}_{i}})$. The task of the visual encoding model is to optimize the image feature extraction model and the voxel mapping model to minimize the MES loss function \[{{L}_{MSE}}=(1/N)\times\sum\nolimits_{i=1}^{N}{{{({{v}_{i}}-{{{\hat{v}}}_{i}})}^{2}}}\], where ${{\hat{v}}_{i}}$ represents the actual fMRI voxel values for the $i$-th stimulating image.

To enhance the visual encoding model using textual descriptions of stimulating images, we need to extract the textual features of these images. Subsequently, aligning the textual feature information with the image feature information via constraints in the loss function is necessary. Therefore, our initial step involves sequencing the textual information of the images and embedding the textual information into a specialized feature space. We use $f_{i}^{text}=F(tex{{t}_{i}})$ to represent the process in which the textual description information of the $i$-th stimulating image is embedded into feature space $f_{i}^{text}$ via model $F^{text}$. In our experiments, the text encoding model $F^{text}$ refers to the pre-trained CLIP model. The pre-trained CLIP has undergone extensive pre-training from various large-scale datasets, possessing exceptional capabilities to extract detailed semantic information from textual inputs.

For the encoded textual information of the $i$-th stimulating image already embedded into the feature space $f_{i}^{text}$, we assume alignment between this feature and the ${{f}_{i}}$ feature of the stimulating image. This assumption allows us to maximize the utilization of the supervision from the LLM model.

To achieve alignment between features $f_{i}^{text}$ and ${{f}_{i}}$, we employ the minimization of a loss function between these two feature sets. Specifically, we reference the infoNCE (Information Noise Contrastive Estimation) from contrastive learning to define our loss function for aligning the image and text features. Our alignment loss function is \[{{L}_{alignment}}=-\log \frac{\exp (f_{i}^{text}\centerdot W{{f}_{i}}/\tau )}{\sum\limits_{i=0}^{N}{\exp (f_{i}^{text}\centerdot W{{f}_{i}}/\tau )}}\], where $W$ represents a learnable parameter matrix aiding in aligning textual features and image features, and $\tau$ is a temperature hyper-parameter used to control the distinctiveness between different samples within the same batch.

So, the final loss function utilized in our laboratory is: \[L = L_{MSE} + \lambda \times L_{alignment}\], where \(\lambda\) is a weight coefficient determining the importance of \(L_{alignment}\) in the total loss function. Leveraging the high-quality image-text descriptions generated via pre-trained LLM, our visual encoding model not only acquires features of stimulating images but also captures high-quality semantic textual information.

\section{Implementation Details}

\subsection{Dataset Preprocessing}

The dataset we used in our experiment is the Natural Scenes Dataset (NSD), which measured high-resolution fMRI responses for tens of thousands of richly annotated natural scenes while participants engaged in continuous recognition tasks. NSD is an extensive collection composed of responses from 8 participants recorded using high-quality 7T fMRI. They were exposed to approximately 73,000 distinct natural scenes to establish a brain encoding model for visual cognition. Across participants 1 to 8, there were 9,841, 9,841, 9,082, 8,779, 9,841, 9,082, 9,841, and 8,779 unique images, respectively. Additionally, corresponding fMRI visual responses were provided for each image. These signals consisted of left hemisphere (LH) and right hemisphere (RH) with 19,004 and 20,544 vertices, respectively, except for participants 6 and 8. Participant 6 had 18,978 LH vertices and 20,220 RH vertices, while participant 8 had 18,981 LH vertices and 20,530 RH vertices \cite{31, nguyen2023algonauts}.

The previous study \cite{gifford2023algonauts} indicates that the visual cortex is divided into various regions with distinct functional characteristics, referred to here as Regions of Interest (ROI). The study provides ROI-level labels for specific areas of the brain, including early visual areas (V1v, V1d, V2v, V2d, V3v, V3d, and hV4), body selective areas (EBA, FBA-1, FBA-2, and mTL-bodies), face selective areas (OFA, FFA-1, FFA-2, mTL-faces, and aTL-faces), place selective areas (OPA, PPA, RSC), and word selective areas (OWFA, VWFA-1, VWFA-2, mfs-words, and mTL-words). Additionally, vertices are assigned to different streams using the stream level mapping: early, midventral, midlateral, midparietal, ventral, lateral, parietal, and unknown. The "unknown" label is applied to vertices that do not belong to any specific ROI or stream in the mappings.

We divided the data from the training set into training, validation, and test sets with proportions of 85\%, 10\%, and 5\%, respectively. In terms of data augmentation, we employed significant resizing and cropping of images to the model's input size (224×224) using a large crop ratio (e.g., 0.8) and square aspect, simulating slight variations in the subject's gaze point. Other augmentations did not yield any benefits and, in some cases, even harmed performance (such as horizontal flipping). For evaluation purposes, only image resizing was adjusted. The fMRI activity targets for each subject were mapped to a common Region of Interest (ROI) space, defined as the union of ROIs masks challenging for all individual subject ROIs. Missing vertices for individual subjects were padded with zeros.

\subsection{Evaluation Metric}

To better assess the encoding performance of our model (i.e., how accurately the model encodes brain responses), we compare the predictions of our model to the empirically measured brain responses. Specifically, we first correlate the predicted test set fMRI with the corresponding real fMRI. Then, we square the correlation coefficients of each vertex, and finally, normalize the resulting value of each vertex by its noise ceiling. The specific formula is as follows \[m=\frac{1}{n}\sum\limits_{i}^{n}{\frac{{{R}^{2}}}{N{{C}_{i}}}}\], where $NC$ (noise ceiling) represents the upper limit of correlation achievable given the noise level in the data \cite{37}.

\subsection{Model Settings}

In our experiments, we utilized the EVA02 model as our core model responsible for extracting stimulating image features, aiming to achieve superior encoding performance. Specially, we used the "eva02\_base\_patch14\_224.mim\_in22k" model in Pytorch library (timm) (image size = 224, patch size=17, embed dim=768, depth=12).

\begin{figure*}[t]
	\centering
	\includegraphics[width=1.0\textwidth]{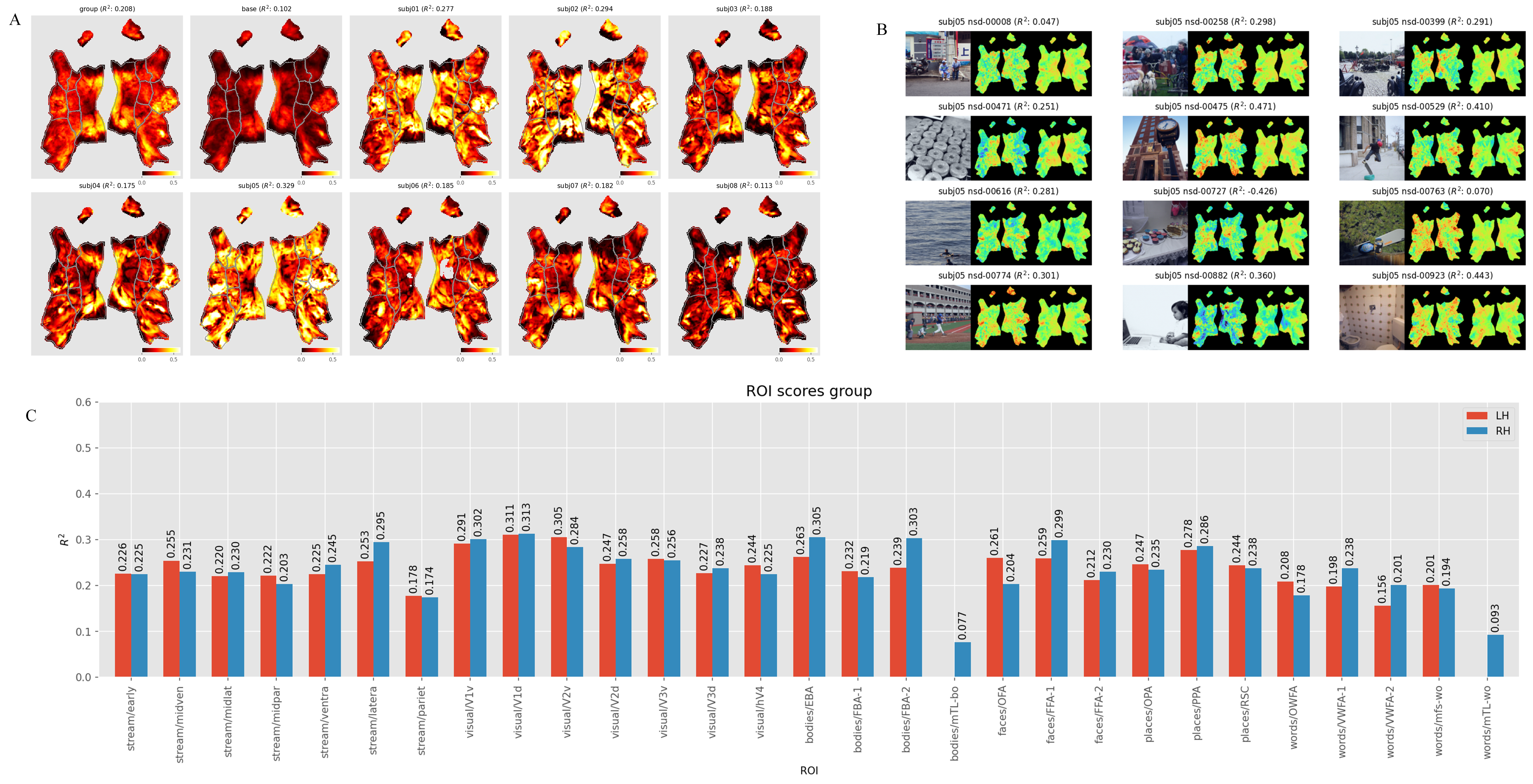} 
	\caption{Prediction performance. \textbf{A} Vertex-wise ${{R}^{2}}$ maps for each subject on the validation split. \textbf{B} Individual sample images (left), fMRI targets (middle), and predictions (right) for NSD Subject 5. \textbf{C} Group median ${{R}^{2}}$ scores for individual ROIs.}
	\label{fig2}
\end{figure*}

\begin{figure*}[!h]
	\centering
	\includegraphics[width=1.0\textwidth]{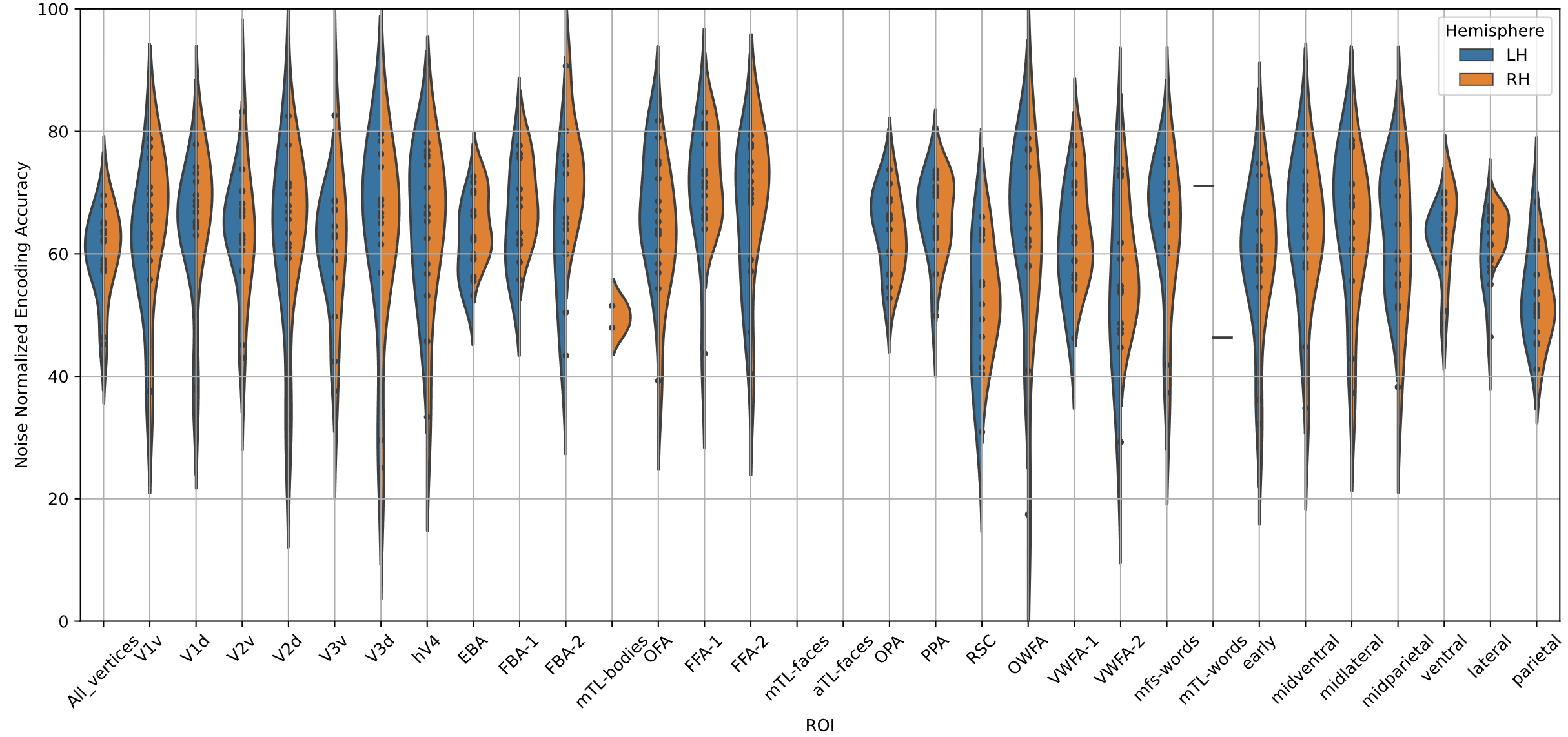} 
	\caption{Performance results of our model in The Algonauts Project 2023 competition.}
	\label{fig3}
\end{figure*}

Our training process is primarily divided into two stages. In the first stage, we freeze the EVA model and select the outputs of layers 0, 2, 4, 6, 8, and 10 of the EVA \cite{fang2023eva} model as the features for stimulating images. In the second stage, we freeze certain components of the EVA model while fine-tuning the other components. Details regarding the training parameters for both stages and the names of frozen components in the EVA model are provided in Tab.1 and Tab. 2.

\begin{table}
	\caption{The training parameters for the \textbf{first stage}}
	\centering
	\begin{tabular}{cc}
		\toprule Parameter & Value\\
		\midrule
		epochs & 40 \\
		batch size & 512 \\
		learning rate & 6.0e-4 \\
		weight decay & 0.8  \\
		dropout & 0.9 \\
		\bottomrule
	\end{tabular}
\end{table}

\begin{table}
	\caption{The training parameters for the \textbf{second stage}}
	\centering
	\begin{tabular}{cc}
		\toprule Parameter & Value\\
		\midrule
		epochs & 6 \\
		batch size & 184 \\
		learning rate & 1.0e-5 \\
		weight decay & 0.8  \\
		dropout & 0.9 \\
		freezen blocks(EVA) & \makecell[c]{cls\_token, pos\_embed\\patch\_embed*, blocks*norm*\\blocks*mlp*} \\ 
		\bottomrule
	\end{tabular}
\end{table}

\section{Experimental Results}

Fig. 2 shows the overall performance of our model. Our team, zcboluo, achieved a score of 58.84 in the first stage and a score of 60.2099 in the second stage of The Algonauts Project 2023 Challenge Competition. Our team secured the 6th position in the competition standings. For more details and results, you can visit the competition website at: https://codalab.lisn.upsaclay.fr/competitions/9304\#results.

\begin{table}
	\caption{The results of the ablation study}
	\centering
	\begin{tabular}{cccc}
		\toprule Model & ${{R}^{2}}$ Value & Score & Inprovement (\%)\\
		\midrule
		stage-1 & 0.204 & 58.83 & - \\
		stage-2 & 0.208112 & 60.18 & - \\
		\makecell{stage-2 \\ ($\lambda =1$)} & 0.194580 & 56.65 & 6.5$\downarrow$ \\
		\makecell{stage-2 \\ ($\lambda =0.1$)}& 0.204344 & 59.13 & 1.8$\downarrow$\\
		\makecell{stage-2 \\ ($\lambda =1e-2$)} & 0.208440 & 60.17 & \textbf{0.15}$\uparrow$ \\
		\makecell{stage-2 \\ ($\lambda =1e-3$)} & 0.208896 & 60.21 & \textbf{0.37}$\uparrow$ \\
		\bottomrule
	\end{tabular}
\end{table}

In Fig. 2A, we conducted visual analysis for each participant. It's evident from the graph that the varying number of experimental sessions completed by participants from different groups led to varying prediction results among subjects. Participant 5 showcased the best performance. Moving to Fig. 2B, we randomly selected several stimulus image predictions from Participant 5, encompassing a range of predictions, including both high and low scores. These samples cover the prediction spectrum within Participant 5. From these samples, it's noticeable that stimulus images with prominent subjects tend to yield better predictions. For instance, in "nsd-00475," where a clock stands out as a highly prominent subject, the prediction score is generally higher. Conversely, in the image "nsd-00727" featuring multiple scattered cakes, the subjects are more dispersed, resulting in lower prediction scores. Fig. 2C illustrates the diverse ${{R}^{2}}$ score performances of all participants' stimulus images across different ROIs. This visualization highlights that our model performs better in lower-level visual areas, such as V1. Fig. 3 presents the performance of our model on the competition website, which aligns with the results we derived from the analysis mentioned above.

\section{Ablation Study}
$\lambda $, the weight coefficient in the loss function $L$, represents the proportional strength between ${{L}_{mse}}$ and ${{L}_{alignment}}$. When the value of $\lambda $ is too large, it causes the textual feature information to dominate as a prominent feature within $L$, subsequently reducing the effectiveness of the encoding performance. As indicated in Tab. 3, $\lambda =1e-3$ consistently achieves the best performance.

\section{Conclusion}

This paper introduces a new multi-modal training paradigm, aligning with LLM, for encoding fMRI activity in visual cortex. This paradigm involves generating high-quality textual descriptions using LLM and aligning stimulus image features with text features by minimizing a contrastive  learning loss function. By extending the original uni-modal feature to the multi-modal feature, this training paradigm enhances the performance of the encoding model. Building upon this paradigm, the paper proposes a model called LLM-VEM. Empowered by the aforementioned training paradigm, LLM-VEM integrates stimulus images and textual descriptions, aligning them to obtain multi-modal feature information and achieving strong performance. In future research, we will explore more possibilities of applying LLM to the field of visual information encoding.

\section{Acknowledgments}

This research was supported by the National Natural Science Foundation of China under grant 62106285.

\bigskip

\bibliography{aaai24}

\end{document}